# Training Physics-Informed Neural Networks via Multi-Task Optimization for Traffic Density Prediction

Bo Wang
*Dept. of Computing Technologies*
*Swinburne University of Technology*
Melbourne, Australia
bwang@swin.edu.au

A. K. Qin
*Dept. of Computing Technologies*
*Swinburne University of Technology*
Melbourne, Australia
kqin@swin.edu.au

Sajjad Shafiei
*Dept. of Computing Technologies*
*Swinburne University of Technology*
Melbourne, Australia
sshafiei@swin.edu.au

Hussein Dia
*Dept. of Civil and Construction Engineering*
*Swinburne University of Technology*
Melbourne, Australia
hdia@swin.edu.au

Adriana-Simona Mihaita
*Data Science Institute*
*University of Technology Sydney*
Sydney, Australia
adriana-simona.mihaita@uts.edu.au

Hanna Grzybowska
*Simulation Group,*
*Data 61|CSIRO*
Sydney, Australia
hanna.grzybowska@data61.csiro.au

*Abstract*— Physics-informed neural networks (PINNs) are a newly emerging research frontier in machine learning, which incorporate certain physical laws that govern a given data set, e.g., those described by partial differential equations (PDEs), into the training of the neural network (NN) based on such a data set. In PINNs, the NN acts as the solution approximator for the PDE while the PDE acts as the prior knowledge to guide the NN training, leading to the desired generalization performance of the NN when facing the limited availability of training data. However, training PINNs is a non-trivial task largely due to the complexity of the loss composed of both NN and physical law parts. In this work, we propose a new PINN training framework based on the multi-task optimization (MTO) paradigm. Under this framework, multiple auxiliary tasks are created and solved together with the given (main) task, where the useful knowledge from solving one task is transferred in an adaptive mode to assist in solving some other tasks, aiming to uplift the performance of solving the main task. We implement the proposed framework and apply it to train the PINN for addressing the traffic density prediction problem. Experimental results demonstrate that our proposed training framework leads to significant performance improvement in comparison to the traditional way of training the PINN.

*Keywords—Multi-task optimization, MTO, Physics-informed neural network, PINN, Knowledge transfer*

I. INTRODUCTION

The physical laws that govern the dynamics of a system can often be described as partial differential equations (PDEs). The solution to such PDEs allows the estimate of the system state over time in the future. However, it is often challenging to solve PDEs due to lack of analytical solutions. Compared to traditional numerical solvers, PINNs [1] provide a new and more efficient way of solving the PDE by approximating the solution to the PDE via the neural network (NN) model while incorporating certain PDE-related regularization terms as the prior knowledge into the NN training. In this way, the training of the NN can be guided by the physical laws which govern the generation of the training data to avoid the need of a large amount of training data by the NN to achieve the desired generalization performance [2].

In recent years, PINNs have been successfully applied in a variety of scientific and engineering scenarios, such as fluid dynamics [3][4] and material engineering [5][6], which show superiority over traditional PDE solvers. However, it is a more challenging task for training PINNs than training traditional NNs. Specifically, the loss function in the PINN is composed of two parts, i.e., the NN loss (related to NN training errors) and the PDE loss (related to PDE residuals), which may result in inconsistent gradient directions for reducing each part in different training stages. Because of the inherent magnitude difference of these two parts and the variability of the training data (for the NN loss) and the sampling data (for the PDE loss) in terms of both quantity and quality, decreasing one part during training is likely to lead to increase of another part. It makes traditional gradient descent-based training techniques very difficult (highly time-consuming) to converge.

Training NNs, particularly deep NNs (DNNs) [7], is by no means trivial, which has attracted considerable research efforts from the past till now. Among existing works, those based on transfer learning have demonstrated excellence. Some of such approaches [8][9] make use of the knowledge in certain forms acquired from pre-trained models to assist in a given training task. Others [10] create multiple auxiliary training tasks which are somewhat relevant to a given training task, and solve them together with the given (main) task while enabling knowledge transfer across different task-solving processes such that the given task can be better solved. In the light of the latter kind of approaches, we propose a novel MTO-based PINN training framework, where auxiliary PINN training tasks are created and solved together with the main training task while enabling cross-task knowledge transfer to improve on the performance of solving the main task.

In our proposed MTO-based PINN training framework, the auxiliary PINN training tasks can be designed from the aspects of using different training data and solving different tasks related to the main task, among others. The PINN models utilized in multiple different tasks share the same architecture except for the front (input) and/or end (output) layers, which can accommodate the data and/or tasks of different properties. Multiple PINNs are individually trained at the same time for solving their own tasks during which network parameters from the other PINNs can be used to help train any specific PINN via a linear combination with learnable coefficients if a certain knowledge transfer criterion is met.

This work is supported by the Australian Research Council (ARC) under Grant No. LP180100114 and DP200102611.

We implement the proposed training framework and apply it to train the PINN model for traffic density prediction which is a promising application of PINNs, where auxiliary tasks are designed by considering both different training data and other different but relevant traffic state estimation (prediction) tasks. Experimental results show that our method can consistently improve over the traditional way of training the PINN across various kinds of auxiliary tasks.

## II. BACKGROUND

### A. PINNs in Traffic State Estimation

Traffic state estimation has gained much research attention in the field of transport engineering because it directly impacts many downstream applications, e.g., traffic control, incident management, and vehicle navigation. In recent years, traffic state estimation based on PINNs have been studied [11][12]. Most of such works utilize macroscopic traffic flow models to depict the physical laws in the PINN, which analyze the traffic in an aggregative manner without considering individual vehicles. The macroscopic models are based on the kinematic wave theory and typically treat a traffic flow as a fluid stream [13][14]. They can capture forward and backward shockwaves that occur when the traffic state varies through a section of the road. Several traffic flow models, such as Lighthill-Whitham-Richards (LWR) [13][14] and Aw-Rascle-Zhang (ARZ) [15], have been adopted to define the PDE component in the PINN. As for the NN component in the PINN, most of the existing works on PINN-based traffic state estimation employ a similar NN architecture as that employed in [2], i.e., one input layer, several hidden layers, and one output layer, where the "Tanh" activation function is usually utilized in the hidden layers. Compared to the NN model, the PINN model can significantly improve traffic prediction performance in the face of a limited number of training data [11][12].

### B. MTO-based NN Training

MTO [16][17] is a newly emerging research frontier in the field of optimization. It allows multiple relevant optimization tasks to be solved simultaneously and the knowledge acquired from solving one task is reused to help solve some other tasks via knowledge transfer so that the performance of solving each individual task gets improved.

In recent years, many new MTO techniques [18][19][20] have been proposed and successfully applied in various fields [21]. One of the promising MTO application scenarios is to train NNs. It is well-known that training an NN corresponds to solve a complex non-convex optimization problem, prone to getting stuck into inferior local optima. To address this issue, the MTO-based NN training first creates multiple auxiliary training tasks relevant to the main training task. Then, both the main and auxiliary tasks are solved at the same time while allowing the knowledge obtained from one task-solving process, often in the form of promising network parameters, to assist in some other task-solving processes via knowledge transfer and reuse. As a result, the main training task can be better solved because the knowledge from relevant auxiliary tasks can help the main training task to jump out of its own inferior local optima. There exist different ways of creating auxiliary training tasks and the knowledge can be transferred and reused in different manners [10][22][23].

## III. THE PROPOSED METHOD

The MTO-based training paradigm for DNNs [10][22] has shown performance superiority over the traditional methods of training DNNs, where multiple auxiliary training tasks related to the given (main) training task are created and then solved together with the main task so that the useful knowledge from any individual task-solving process, e.g., network parameters, can be transferred and used to help solve some other training tasks, aiming to improve the performance of solving the main task. We extend this idea for training PINNs and choose traffic state prediction as the application scenario, where different traffic state variables that are somewhat related and multiple traffic data sets in different spatial and time zones are utilized for designing auxiliary tasks, relevant but different from the main task, to enable MTO-based training. In what follows, we will first introduce PINN's formulation in traffic state prediction and then elaborate the MTO-based PINN training framework.

### A. Traffic PINNs

Suppose the task is to estimate traffic state $u$ (e.g., traffic flow $q$, density $k$, and speed $v$.) over time $t$ within a time span of $T$ (seconds) in the future along a road segment of length $D$ (meters). The PINN model, as illustrated in Fig. 1 will take as inputs time $t$ and distance $d$ (with respect to the road origin) and output $\hat{u}$. Different from the traditional NN, PINN's loss function $Loss$ has two parts, i.e., $Loss_{NN}$ and $Loss_{PDE}$.

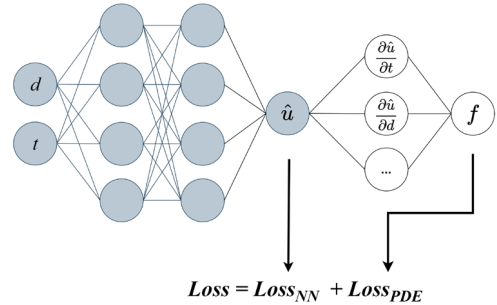

Fig. 1. A simple illustration of the PINN model.

$Loss_{NN}$ is typically formulated as the mean squared error (MSE) between NN's output $\hat{u}$ and the corresponding ground truth $u$ as follows:

$$Loss_{MSE} = \frac{1}{n_u}\sum_{i=1}^{n_u}|\hat{u}(d_i^u,t_i^u) - u_i^u|^2 \quad (1)$$

where $M_u = \{d_i^u, t_i^u, u_i^u | i = 1, \dots n_u\}$ represents a training set with $n_u$ available training samples.

The definition of $Loss_{PDE}$ depends on the adopted PDE that governs traffic dynamics. Suppose the residual function of the PDE is denoted as $f(d, t, u)$. $Loss_{PDE}$ is formulated as:

$$Loss_{PDE} = \frac{1}{n_f}\sum_{i=1}^{n_f}\left|f\left(d_i^f, t_i^f, \hat{u}_i^f\right)\right|^2 \quad (2)$$

where $M_f = \{d_i^f, t_i^f, \hat{u}_i^f | i = 1, \dots n_f\}$ is a set of $n_f$ randomly sampled $d$ and $t$ from $[0, D]$ and $[0, T]$, respectively and the corresponding NN's outputs when they are fed into the NN as inputs.

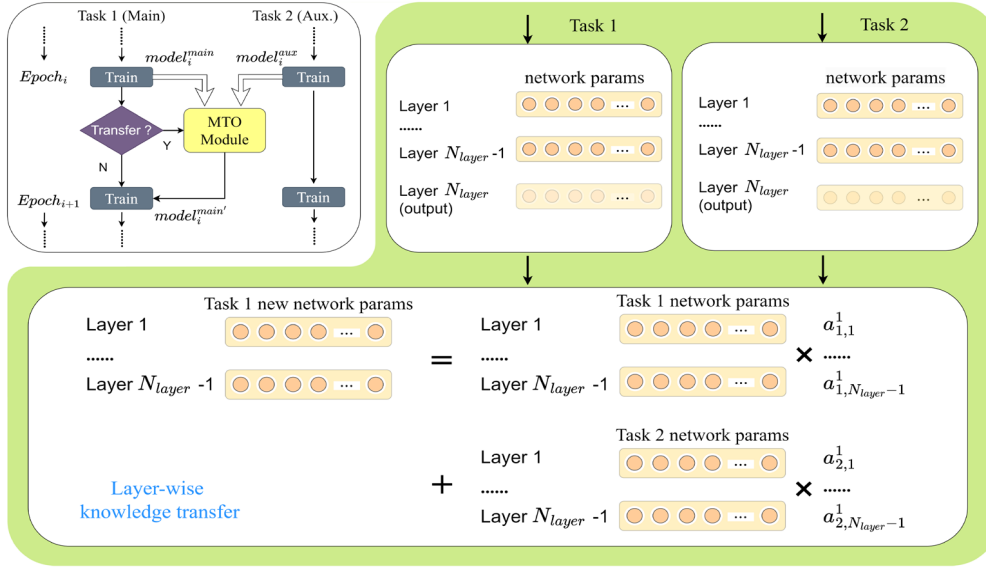

Fig. 2. The basic working principle of the proposed MTO-based training framework.

Following previous studies [13][14][24], we will employ Greenshields's LWR traffic flow model which can provide a macroscopic way for describing the evolution of traffic density $k$ on a road network. It is based on an assumption that traffic density is a continuous function with respect to space and time, where traffic flow $q$ is defined as the product of traffic density $k$ and local traffic speed $v$ [13][14]. Th model is underpinned by the following PDE:

$$\frac{\partial q}{\partial d} + \frac{\partial k}{\partial t} = 0 \quad (3)$$

$$q = k \times v \quad (4)$$

$$v = v_f\left(1 - \frac{k}{k_j}\right) \quad (5)$$

where $v_f$ and $k_j$ represent the free flow speed and the jammed density, respectively. From this PDE, we define the residual function with respect to density $k$ or speed $v$ as follows:

$$f(d,t,k) = v_f \frac{\partial k}{\partial d} - 2\frac{v_f}{k_j}\frac{\partial k}{\partial d}k + \frac{\partial k}{\partial t} \quad (6)$$

$$f(d,t,v) = \frac{\partial v}{\partial d}k_j - 2\frac{k_j}{v_f}\frac{\partial v}{\partial d}v - \frac{k_j}{v_f}\frac{\partial v}{\partial t} \quad (7)$$

Depending on the considered prediction task (density or speed), one of the above will be used as $Loss_{PDE}$ in the PINN. Therefore, the overall loss of the PINN is defined as:

$$Loss = Loss_{NN} + Loss_{PDE} \quad (8)$$

By combining the PDE-based traffic flow model and the data-driven NN model, the PINN model may rely on merely a limited amount of training data to obtain accurate prediction of traffic states. It aligns well with real-world scenarios where traffic sensors are often sparsely deployed.

*B. MTO-based PINN Training*

Our proposed MTO-based PINN training framework starts with the design of auxiliary training tasks relevant to the main training task. In this work, the main task is defined as training a PINN for traffic density prediction based on a training set with known traffic density. Considering that traffic density and speed are inherently related, auxiliary tasks can be designed as training a PINN with the exact same architecture for traffic density prediction based on a different training set with known traffic density, training a PINN with the nearly same architecture (except for the output layer) for traffic speed prediction based on the same training set with known traffic speed, and training a PINN with the nearly same architecture (except for the output layer) for traffic speed prediction based on a different training set with known traffic speed, among others.

The main and auxiliary training tasks will be individually solved at the same time. In the process of solving any task, when a certain knowledge transfer criterion is met, an MTO module will be triggered for incorporating the knowledge (in the form of network parameters) acquired from the other tasks. The training process for any task will be terminated when the pre-defined maximum number of training epochs is reached. In this work, we adopt an adaptive MTO triggering strategy. Specifically, if the best training loss value obtained so far cannot be improved by a decent amount (e.g., a pre-defined percentage of the best loss value obtained so far) over a period of S consecutive training epochs, so-called MTO triggering window size, the MTO module will be triggered.

Suppose the MTO module is triggered for Task $k$ ($k \in \{1, ..., N_{task}\}$) at epoch $i$. The current network parameters from Task $k$ will be linearly combined with those from the other $N_{task} - 1$ tasks in a layer-wise manner to generate the new network parameters for Task $k$. The combination coefficients $\boldsymbol{\alpha}^k = \{a_{ij}^k | i = 1, ..., N_{task}, j = 1, ..., N_{layer} - 1, a_{ij}^k \in R\}$ will be learned via solving Task $k$ with respect to $\boldsymbol{\alpha}^k$ while the original network parameters from all tasks are kept frozen. As such, the knowledge from the other tasks can be adaptively transferred into Task $k$. Notably, the last (output) layer of the NN is not involved in the knowledge transfer process because the parameters in that layer are very task-specific and thus less suitable for cross-task knowledge transfer. After $\boldsymbol{\alpha}^k$ is learned for a pre-specified number of training epochs, the new network parameters created upon it will be compared with the original (pre-MTO) network parameters in terms of the corresponding training loss values. The network parameters with the smaller loss value will enter epoch $i+1$, and meanwhile the period of S consecutive training epochs will be re-counted from scratch.

TABLE I. COMPARISON OF THE PROPOSED TRAINING METHOD USING EACH OF THE THREE DIFFERENCE AUXILIARY TASKS WITH THE TRADITIONAL TRAINING METHOD ON TWO TRAINING SETS A AND B IN TERMS OF THE MEAN AND STANDARD DEIVATION OF THE LOSS VALUES ON THE TRAINING SET AND THE MAPE VALUES ON THE TEST SET OVER 10 RUNS. THE STATISTICAL T-TEST AT THE SIGNIFICANCE LEVEL OF 0.05 IS PERFORMED TO COMPARE THE PROPOSED AND TRADITONAL METHODS WITH THE PROPOSED METHOD HIGHLIGHTED IN BOLD IF ITS PERFORMANCE IS STASITICALLY BETTER.

| Method | Main Task Task (Data) | Main Task PDE | Auxiliary Task Task (Data) | Auxiliary Task PDE | Training Set Loss | Test Set MAPE |
|---|---|---|---|---|---|---|
| NN | Density (A) | -- | -- | -- | 11.385 ± 0.612 | 1.279 ± 0.058 |
| PINN | Density (A) | LWR | -- | -- | 14.734 ± 0.647 | 0.357 ± 0.004 |
| PINN+MTO | Density (A) | LWR | Density (B) | LWR | **12.958 ± 0.769** | **0.252 ± 0.039** |
| PINN+MTO | Density (A) | LWR | Speed (A) | LWR | **12.936 ± 0.981** | **0.247 ± 0.017** |
| PINN+MTO | Density (A) | LWR | Speed (B) | LWR | **12.486 ± 0.698** | **0.260 ± 0.033** |
| NN | Density (B) | -- | -- | -- | 10.479 ± 0.930 | 0.298 ± 0.046 |
| PINN | Density (B) | LWR | -- | -- | 15.971 ± 0.881 | 0.171 ± 0.004 |
| PINN+MTO | Density (B) | LWR | Density (A) | LWR | **15.181 ± 0.289** | **0.169 ± 0.002** |
| PINN+MTO | Density (B) | LWR | Speed (A) | LWR | **14.882 ± 0.303** | **0.166 ± 0.001** |
| PINN+MTO | Density (B) | LWR | Speed (B) | LWR | **15.071 ± 0.948** | **0.167 ± 0.003** |

In this work, we choose to employ only one auxiliary task for simplicity. Fig. 2 illustrates the basic working principle of the proposed framework with $N_{task}=2$ as well as the involved MTO module.

## IV. EXPERIMENTS

### A. Data Description

We employ the US Highway 101 data set from the Next Generation Simulation (NGSIM) program for our study. The area of study for data collection spans around 640 meters and comprises five main lanes. The data set contains traffic data captured during a 45-min period in the morning peak, where the raw data have been aggregated every five seconds. We utilize the traffic density, speed, and flow information in the study area and exclude the initial and final road sections from our analysis due to incomplete data.

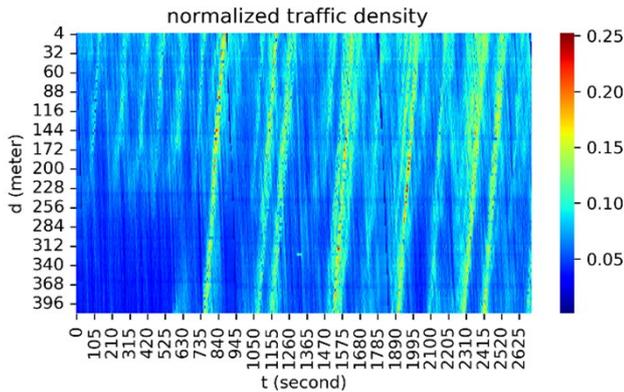

Fig. 3. The spatiotemporal distribution of normalized traffic density for the area in the US 101 data set which is considered in this work.

Fig. 3 illustrates the distribution of normalized traffic density (via the min and max density values in the entire study area) with respect to distance $d$ (from the road origin) and time $t$ (in the 45-min period) for the area in the US 101 data set which is considered in this work. It illustrates multiple shockwaves, indicating the traffic congestion propagation on the freeway. We assume that in practice traffic data are only available in a few locations on the road due to the sparsely deployed traffic sensors. To evaluate the utility of the PINN in this scenario, we create two training sets from the whole data set. Data set A comprises the data from sensors located at $d = 4, 8, 12, 16, 20,$ and 24 meters, respectively at time stamps in [0, 5, 10, …, 2695], and data set B contains the data from sensors located at $d = 8, 92, 128, 292,$ and 408 meters, respectively at time stamps in [0, 5, 10, …, 2695]. In contrast, data set B covers a wider study area than A. In the two training sets, both distance and time are employed as the inputs, and either traffic density or speed can be adopted as the output. The test set for measuring the prediction performance of the PINN contains the data from sensors located at $d \in [4, 8, 12, …, 408]$, respectively at time stamps in [0, 5, 10, …, 2695].

### B. Task Design

The **main task** in this work is defined as training a PINN for making traffic density prediction based on a training set with known traffic density, i.e., Density (A) or Density (B).

The **auxiliary tasks** in this work are designed to be one of the following three training tasks:

- training a PINN with the same architecture as that used in the main task for traffic density prediction but based on a different training set (from the main task) with known traffic density, i.e., Density (B) or Density (A).
- training a PINN with the nearly same architecture (except for the output layer) as that used in the main task for traffic speed prediction and based on the same training set (as the main task) with known traffic speed, i.e., Speed (A) or Speed (B), and
- training a PINN with the nearly same architecture (except for the output layer) as that used in the main task for traffic speed prediction but based on a different training set (from the main task) with known traffic speed, i.e., Speed (B) or Speed (A).

### C. Experimental Settings

The NN component of all the PINN models used in our experiments adopts the same architecture as that employed in [11][12], i.e., a fully connected feedforward NN with one input, eight hidden, and one output layers. For training, we set batch size as 1024, and employ the Lamb optimizer [25] for training both network parameters and layer-wise weights (in the MTO module). The maximum number of training epochs for training the PINN is set to 4,000 and the MTO-triggering window size $S$ is set to 50 with the improvement percentage set to 0.01 (of the best training loss obtained so far). Also, the maximum number of training epochs for learning layer-wise weights in MTO is set to 200. Each method in comparison is executed for 10 independent runs.

For performance evaluation, we employ the training loss of the PINN model to evaluate the training performance, and the mean absolute percentage error (MAPE) on the test set to measure PINN's generalization performance. We perform the statistical t-test at the significance level of 0.05 to compare the 10-run results of any two methods under consideration.

*D. Overall Comparision*

We compare the proposed PINN training method with the traditional PINN training method for learning the same PINN model (for traffic density prediction) by using training sets A and B, respectively. Therefore, the main task is either Density (A) or Density (B). The auxiliary task adopted in the proposed training method is Density (B), Speed (A) or Speed (B) with respect to main task Density (A), and Density (A), Speed (A) or Speed (B) with respect to main task Density (B). Further, we perform a comparison with the NN-only counterpart of the PINN model to reveal the utility of the PINN.

The experimental results are reported in Table I. It can be observed that the proposed MTO-based PINN training method consistently outperforms the traditional PINN training method in terms of both the loss on the training set and the MAPE on the test set for all three auxiliary tasks on both training sets A and B. Among three adopted auxiliary tasks, the speed task on a different training set outperforms the other two in terms of the eventually obtained training loss on both training sets. In addition, although the NN model can be trained to achieve the much better training loss, the corresponding MAPE on the test set is worse, implying overfitting and thus poor generalization.

*E. Ablation Study*

  *a) MTO Triggering Strategy*

In this work, we adopt an adaptive strategy for triggering the MTO module, where the MTO is triggered whenever the best training loss value obtained so far cannot be improved by 1% over $S$ consecutive epochs. To validate its effectiveness, we compare it with a fixed triggering strategy which regularly triggers the MTO module every 50 training epochs. Table II reports the comparison results of these two strategies in terms of their eventually obtained training loss values with respect to three different auxiliary tasks and two different training sets. We perform the statistical t-test at the significance level of 0.05 to compare these two strategies and highlight in bold the best one (with the minimum mean loss value over 10 runs) and also another if it is statistically similar to the best. It can be observed that the adaptive strategy consistently performs best across all the comparison scenarios. It is noteworthy that the adaptive strategy typically triggers the MTO module less often than the fixed strategy when using the same MTO triggering window size. For example, when the triggering window size is set to 50 and the total number of training epochs is set to 4,000, as employed in this work, the MTO triggering times for adaptive and fixed strategies are 60 and 80, respectively. It implies that the adaptive strategy may potentially reduce the computational cost by reducing MTO execution times.

Fig. 4 illustrates the convergence curves of the training loss with respect to the proposed training methods equipped with the adaptive and fixed MTO triggering strategies and the traditional training method without MTO, respectively. It can be observed that regularly triggering the MTO module may result in the worse performance than the traditional training method. By using the adaptive strategy, MTO is not triggered in the beginning stage of training, but gradually and adaptively triggered to make performance improvement once the training process is stuck, as evidenced by the distribution of triggering points displayed in Fig. 4.

TABLE II. COMPARISON OF THE TRAINING LOSS (MEAN AND STANDARD DEVIATION OVER 10 RUNS) WHEN USING TWO DIFFERENT MTO TRIGGERING STRATEGIES IN THE PROPOSED METHODS WITH DIFFERENT AUXILIARY TASKS ON TWO TRAINING SETS. THE STATISTICAL T-TEST AT THE SIGNIFICANCE LEVEL OF 0.05 IS PERFORMED TO COMPARE THE TWO STRATEGIES WITH THE STATISTICALLY BEST ONE(S) HIGHLIGHTED IN BOLD.

| Main (Data) | Auxiliary (Data) | Fixed | Adaptive |
|---|---|---|---|
| Density (A) | Density (B) | **13.143 ± 0.775** | **12.958 ± 0.769** |
| Density (A) | Speed (A) | 13.473 ± 0.671 | **12.936 ± 0.981** |
| Density (A) | Speed (B) | 13.435 ± 0.776 | **12.486 ± 0.698** |
| Density (B) | Density (A) | **15.258 ± 0.816** | **15.181 ± 0.289** |
| Density (B) | Speed (A) | 15.643 ± 0.382 | **14.882 ± 0.303** |
| Density (B) | Speed (B) | 16.756 ± 1.060 | **15.071 ± 0.948** |

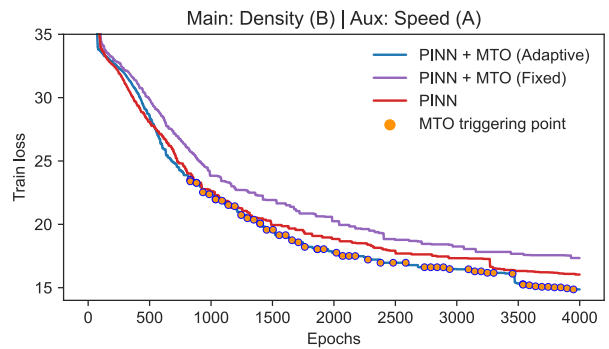

Fig. 4. Training loss convergence curves for the proposed training method with the adaptive or fixed MTO triggering strategy and the traditional training method, where the main and auxiliary tasks are Density (B) and Speed (A), respectively. Training epochs at which MTO is triggered when using the adaptive strategy are denoted by yellow dots.

  *b) Layer-wise Weight Initialization in the MTO Module*

In the MTO module, layer-wise weights $\boldsymbol{\alpha}^k$, $k = 1, \ldots, N_{task}$ are learned based on the training data to enable cross-task knowledge transfer. To investigate the sensitivity of the weight learning to initialization, typically suffered by the NN training, we evaluate several different initial layer-wise weight settings. Suppose $\boldsymbol{\alpha}^1$ and $\boldsymbol{\alpha}^2$ are the main and auxiliary tasks under consideration.

Table III shows the comparison results among five initial weight settings, where the four pairs of numbers refer to the same initial weight value for all the layers in the corresponding task, e.g., (1.0, 0.0) denotes the initial weight value of 1.0 for all layers in the main task and 0.0 for all layers in the auxiliary task; Xavier [7] refers to the uniformly random initialization. It can be observed that initialization sensitivity is not severe, though among the five compared settings, the best performer is (1.0, 0.0). In fact, this setting inherently allows for the fine-tuning of network parameters for the main task by making use of the network parameters from auxiliary tasks, and thus avoid the dramatic parameter update which may lead to performance degradation.

TABLE III. COMPARISION OF THE TRAINING LOSS (MEAN AND STANDARD DEVIATION OVER 10 RUNS) WHEN USING FIVE DIFFERENT WAYS OF INITIALIZING LAYER-WISE WEIGHTS IN THE PROPOSED METHODS WITH DIFFERENT AUXILLARY TASKS ON TWO TRAINING SETS. THE STATISTICAL T-TEST AT THE SIGNIFICANCE LEVEL OF 0.05 IS PERFORMED TO COMPARE THE INITIALIZATION METHOD WITH THE BEST MEAN LOSS VALUE (HIGHLIGHTED IN BOLD) WITH EACH OF THE REST, WHERE THE CO-WINNER(S) ARE ALSO HIGHLIGHTED IN BOLD.

| Main (Data) \| Auxiliary (Data) | ($\alpha^1, \alpha^2$) | | | | |
|---|---|---|---|---|---|
| | (1.0, 0.0) | (0.7, 0.3) | (0.5, 0.5) | (0.0, 1.0) | Xavier [7] |
| Density (A) \| Density (B) | **12.958 ± 0.769** | **13.014 ± 0.787** | **12.432 ± 0.999** | **13.004 ± 0.567** | 13.255 ± 0.459 |
| Density (A) \| Speed (A) | **12.936 ± 0.981** | 13.404 ± 0.495 | 13.349 ± 0.475 | **13.165 ± 0.801** | 13.609 ± 0.919 |
| Density (A) \| Speed (B) | **12.486 ± 0.698** | **12.917 ± 0.806** | 13.603 ± 0.600 | **13.090 ± 1.287** | **12.965 ± 0.957** |
| Density (B) \| Density (A) | 15.181 ± 0.289 | **15.067 ± 0.387** | 15.268 ± 0.617 | 15.530 ± 0.282 | 15.402 ± 0.533 |
| Density (B) \| Speed (A) | **14.882 ± 0.303** | 15.263 ± 0.590 | 16.097 ± 0.484 | 15.767 ± 0.345 | 15.526 ± 0.277 |
| Density (B) \| Speed (B) | **15.071 ± 0.948** | 15.884 ± 0.409 | **15.119 ± 0.345** | 15.703 ± 0.424 | **15.264 ± 0.284** |

## F. Parameter Sensitivity Analysis

As discussed previously, the adaptive MTO triggering strategy plays an essential role in the proposed PINN training framework, where the MTO triggering window size $S$ is a key parameter involved. In this section, we conduct a parameter sensitivity analysis on $S$ with respect to the performance of the proposed training method. Specifically, we evaluate the total number of times that the MTO module gets triggered during the entire training period and the eventually obtained training loss values over 10 runs as the MTO triggering window size varies from 10 to 120 at the step size of 10 for the proposed method equipped with each of the three different auxiliary tasks on training set A. The results are illustrated in Fig. 5, which indicate that when the window size increases, the total number of MTO triggering times decreases and the training performance (in terms of the loss) fluctuates and accordingly the best window sizes, based on the training performance, for different auxiliary tasks are not same. In this work, we choose to use the window size of 50 in all experiments considering the overall performance across different auxiliary tasks.

## V. CONCLUSIONS AND FUTURE WORK

We proposed a novel PINN training framework based on MTO. In this framework, one or more auxiliary training tasks are created and solved together with the main training task so that solving auxiliary tasks may help better solve the main task via knowledge transfer, where an MTO module is designed to enable knowledge transfer. We applied the proposed method to train the PINN for traffic density prediction, where different auxiliary tasks were designed and evaluated. Compared to the traditional PINN training method, our proposed MTO-based method demonstrated the performance advantages in terms of both training and testing. Our future work includes evaluation of the proposed method by using more than one auxiliary tasks and extending its application to the other traffic problems and non-traffic scenarios. We will also study how to better balance the two loss parts in an adaptive manner based on the strategies proposed in our previous works [26][27] to facilitate training.

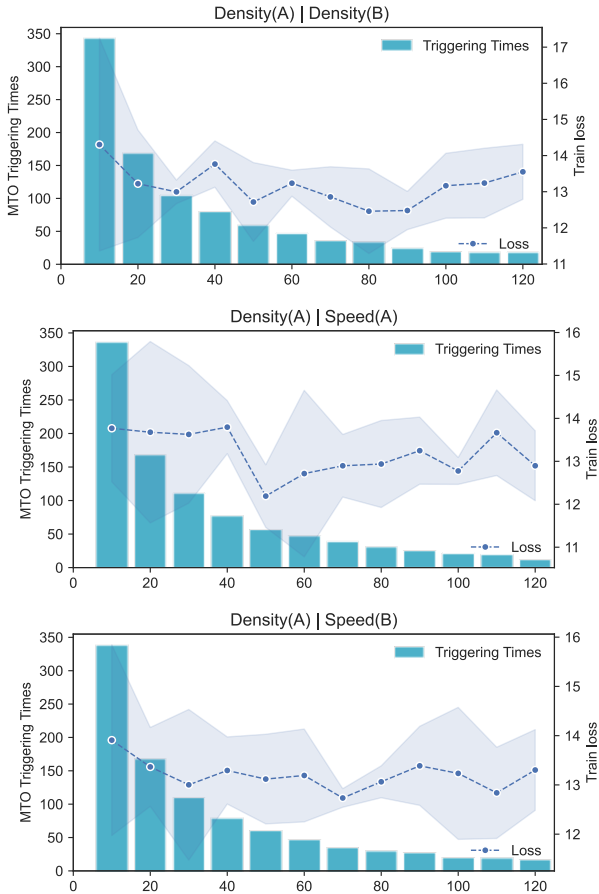

Fig. 5. Comparison of the total number of times for triggering the MTO module during the entire training period and the distribution of training loss values over 10 runs as the MTO triggering window size varies from 10 to 120 at the step size of 10 (horizontal axis), for the proposed training method using each of the three different auxiliary tasks on training set A.